\documentclass[conference]{IEEEtran}
\IEEEoverridecommandlockouts
\usepackage{cite}
\usepackage{amsmath,amssymb,amsfonts}
\usepackage[noend]{algpseudocode}
\usepackage{algorithm}
\usepackage{graphicx}
\usepackage{wrapfig}
\usepackage{textcomp}
\usepackage{hyperref}
\usepackage{tabularx}
\usepackage{array}
\usepackage{color}
\usepackage{multirow}

\def\BibTeX{{\rm B\kern-.05em{\sc i\kern-.025em b}\kern-.08em
    T\kern-.1667em\lower.7ex\hbox{E}\kern-.125emX}}
\begin{document}

\title{``How to rate a video game?'' - A prediction system for video games based on multimodal information}

\author{\IEEEauthorblockN{1\textsuperscript{st} Vishal Batchu}
\IEEEauthorblockA{\textit{Centre for Visual Information Technology} \\
\textit{International Institute of Information Technology Hyderabad}\\
Hyderabad, India \\
vishal.batchu@students.iiit.ac.in}
\and
\IEEEauthorblockN{2\textsuperscript{nd} Varshit Battu}
\IEEEauthorblockA{\textit{Language Technologies Research Center} \\
\textit{International Institute of Information Technology Hyderabad}\\
Hyderabad, India \\
battu.varshit@research.iiit.ac.in}
\and
\IEEEauthorblockN{3\textsuperscript{rd} Murali Krishna Reddy}
\IEEEauthorblockA{\textit{Language Technologies Research Center} \\
\textit{International Institute of Information Technology Hyderabad}\\
Hyderabad, India \\
murali.dakannagari@research.iiit.ac.in}
\and
\IEEEauthorblockN{4\textsuperscript{th} Radhika Mamidi}
\IEEEauthorblockA{\textit{Language Technologies Research Center} \\
\textit{International Institute of Information Technology Hyderabad}\\
Hyderabad, India \\
radhika.mamidi@iiit.ac.in}
}

\maketitle

\begin{abstract}
Video games have become an integral part of most people's lives in recent times. This led to an abundance of data related to video games being shared online. However, this comes with issues such as incorrect ratings, reviews or anything that is being shared. Recommendation systems are powerful tools that help users by providing them with meaningful recommendations. A straightforward approach would be to predict the scores of video games based on other information related to the game. It could be used as a means to validate user-submitted ratings as well as provide recommendations. This work provides a method to predict the G-Score, that defines how good a video game is, from its trailer (video) and summary (text). We first propose models to predict the G-Score based on the trailer alone (unimodal). Later on, we show that considering information from multiple modalities helps the models perform better compared to using information from videos alone. Since we couldn't find any suitable multimodal video game dataset, we created our own dataset named VGD (Video Game Dataset) and provide it along with this work. The approach mentioned here can be generalized to other multimodal datasets such as movie trailers and summaries etc. Towards the end, we talk about the shortcomings of the work and some methods to overcome them.
\end{abstract}


\section{Introduction}
Video games are almost everywhere these days, from individual consumers who play video games for fun to serious E-Sports professionals. The video game industry is a billion dollar industry, it was valued at \$44.9 billion back in 2007 which rose to \$91.5 billion in 2015. The increase in the rate of development of games and the number of people who play these games spiked up hand in hand throughout the world over the recent years. This increase in the sheer number of games marketed required people to rely on a trusted resource that would give them information about these games since it is infeasible for a human to keep the details of every single game ever released in memory. Another trend observed in recent times is that there is an exponential increase in the amount of data shared online. This, however, comes with certain unforeseen consequences such as a reduction in the quality of data present online, the spread of bogus information i.e false information being shared online. Considering the video game industry, people often rely on various sites to provide them with ratings, reviews etc of games before purchase. Since most ratings and reviews are submitted by a wide array of users, maintaining them is hard and hence, we end up having a lot of incorrect/unwanted entries. Another issue we often face with simple methods of input is that users might unknowingly select the wrong option such as an incorrect rating or a genre for a video game. Reviews and descriptions don't face this issue since textual inputs have lesser tendency to be incorrectly entered, however not many people would be willing to spend their time adding textual information and hence we see a wide use of simple input methods. Recommendation systems are quite popular since they allow us to provide meaningful options to users for various purposes. Deep learning has shown a lot of promise at this task. We define the G-Score of a game as a value that determines how good a game is which is derived from critic and user game ratings. In order to mitigate the issues mentioned earlier and to offer useful recommendations to users, we propose several deep neural network architectures that would predict the G-Score from the trailer and the summary of a video game. We believe that the use of summaries along with the trailers would aid the model to predict the G-Score better compared to the use of trailers alone. This would also aid game developers while creating trailers to see how well they score before a public release since the predicted G-Scores could be used to refine and improve the trailers. In order to train our models, we have created the VGD dataset and provide it along with this work.

\section{Related Work}
There have been multiple works in areas related to video and text classification, however, they often deal with domain specific information. Nominal work has been done on video game trailers in the past. We use video game trailers along with reviews to perform a cross-domain analysis in order to predict ratings.

{\bf Video Analysis and Classification} - Zhang et al.\cite{zhang2016video} propose a supervised learning technique for summarizing videos by automatically selecting key-frames. They use Long-Short-Term Memory to model the variable-range temporal dependency among frames so that both representative and compact video summaries can be generated. Venugopalan et al.\cite{venugopalan2016improving} look into how linguistic knowledge taken from large text corpus can aid the generation of natural language descriptions of videos. Haninger et al.\cite{haninger2004content} quantified and characterized the content in video games rated T (for "Teen") and measure how accurate the ESRB-assigned content descriptors displayed on the game box are to the real game. Simonyan et al.\cite{NIPS2014_5353} investigate architectures of discriminatively trained deep Convolutional Networks for action recognition in videos. Capturing the complementary information from still frames and motion between frames was a challenge they address. Kahou et al.\cite{kahou2016emonets} present an approach to learn several specialist models using deep learning techniques. Among these are a convolutional neural network focusing on capturing visual information in detected faces, a deep belief net which focuses on the representation of the audio stream, a K-Means based "bag-of-mouths" model, which extracts visual features around the mouth region and a relational auto-encoder, which addresses spatiotemporal aspects of videos. Le et al.\cite{5995496} present unsupervised feature learning as a way to learn features directly from video data. They presented an extension to the Independent Subspace Analysis algorithm to learn invariant spatiotemporal features from unlabeled video data. Zhou et al.\cite{zhou2007multi} formalize multi-instance multi-label learning in which each training example is associated with not only multiple instances but also multiple class labels. They propose algorithms for scene classification based on the relationship between multi-instance and multi-label learning.

{\bf Text Analysis and Classification} - Glorot et al.\cite{glorot2011domain} propose a deep learning approach that learns to extract a meaningful representation for each review in an unsupervised manner. Sentiment classifiers trained with this high-level feature representation clearly outperform state-of-the-art methods. Zhang et al.\cite{NIPS2015_5782} show empirical exploration on the use of character-level convolutional networks (ConvNets) for text classification. They built large-scale datasets to show that character-level convolutional networks can achieve state-of-the-art results. Iyyer et al.\cite{iyyer2015deep} present a simple deep neural network that competes with and sometimes outperforms models on sentiment analysis and factoid question answering tasks by taking only a fraction of the training time. Baker et al.\cite{baker1998distributional} describe the application of Distributional Clustering to document classification. Their approach clusters words into groups based on the distribution of class labels of each word. Unlike techniques such as Latent Semantic Indexing, they were able to compress the feature space, while maintaining the classification accuracy. Poria et al.\cite{poria2015deep} use the extracted features in multimodal sentiment analysis of short video clips representing one sentence each. They use the combined feature vectors of textual, visual, and audio modalities to train a classifier which is based on multiple kernel learning, which is known to be good at heterogeneous data. Zhang et al.\cite{zhang2009feature} mention that this learning problem is addressed by using a method called M\textsubscript{LNB} which adapts the traditional naive Bayes classifiers to deal with multi-label instances. Feature selection mechanisms are incorporated into M\textsubscript{LNB} to improve its performance.

\section{Dataset}
We have created a dataset named VGD that consists of the trailer, summary, developer, age rating, user-score, critic-score and genre of 1,950 video games. The data was collected from metacritic.com\footnote{\url{http://www.metacritic.com/game}}. The dataset along with the code used can be found at \href{https://goo.gl/Z8bNN3}{https://goo.gl/Z8bNN3} for replicability and future use. This is the first dataset of its kind and we believe it would be quite helpful to the research community.

\begin{table}[H]
\setlength{\extrarowheight}{0.3em}
	\hspace{1cm}
	\begin{minipage}{.1\textwidth}
        \centering
        \begin{tabular}{|c|c|}
          \hline
          \bf{Genre} & \bf{Entries} \\
          \hline
          \bf{Role-Playing} & 522\\
          \bf{Strategy} & 329\\
          \bf{Action} & 734\\
          \bf{Sports} & 200\\
          \bf{Miscellaneous} & 165\\
          \hline
		\end{tabular}
        \rule{0pt}{2.6ex}
         \begin{tabular}{|c|c|}
          \hline
          \bf{Age Rating} & \bf{Entries} \\
          \hline
          \bf{Mature} & 449\\
          \bf{Adults Only} & 2\\
          \bf{Everyone 10+} & 365\\
          \bf{Teen} & 704\\
          \bf{Everyone} & 409\\
          \bf{Other} & 522\\
          \hline
		\end{tabular}
    \end{minipage}
    \hspace{2cm}
    \begin{minipage}{.1\textwidth}
    	\begin{tabular}{|c|c|}
          \hline
          \bf{G-Score (S)} & \bf{Entries} \\
          \hline
          \bf{0-10} & 0\\
          \bf{11-20} & 3\\
          \bf{21-30} & 7\\
          \bf{31-40} & 19\\
          \bf{41-50} & 89\\
          \bf{51-60} & 263\\
          \bf{61-70} & 433\\
          \bf{71-80} & 693\\
          \bf{81-90} & 416\\
          \bf{91-100} & 27\\
          \hline
		\end{tabular}    
	\end{minipage}
    \vspace{0.7cm}
\caption{Distribution of VGD according to Genre classes, G-Score classes and Age Ratings}
\label{table:dataset}
\end{table}

\subsection{Preprocessing}
The first step in preprocessing involved the removal of certain games that had missing details (a lot of games did not have trailers).

{\bf Trailers} - The trailers extracted from the website had a resolution of 720p (with a few exceptions). We reduced the resolution to 360p since 720p required more space and mostly consisted of redundant information from the view of a neural network. We put an upper limit of 3 minutes for each trailer, trimming trailers that were larger to the 3-minute mark.

{\bf Summaries} - We remove non-ASCII characters from the summaries since some of the summaries had terms from other languages like Japanese, Korean, French etc. However, since they are quite small in number, including them would not provide much value in terms of generalizability of the approach.

The final dataset consists of 1,950 video game trailers and summaries. 

\subsection{Statistics}
Various statistics related to the dataset are provided, that show the diversity of the dataset. Video games were collected from a wide range of over 730 developers. Games span across various age ratings from E (Everyone) to M (Mature) which provides us a wide collection of games.

{\bf Genres} -
We cluster the genres into 5 groups based on similarity as specified in Table \ref{table:genreclasses} and present the number of games belonging to each group in sub-tables in Table \ref{table:dataset}.

\begin{table}[h]
\centering
\setlength\tabcolsep{0.5em}
\setlength{\extrarowheight}{0.5em}
\begin{tabular}{|c|c|}
\hline 
{\bf Genre-Class} & {\bf Genres} \\
\hline
Role-Playing & Adventure, First-Person, Third-Person, Role-Playing \\
Strategy & Turn-Based, Strategy, War-Game, Puzzle, Platformer \\
Action & Action \\
Sports & Fighting, Sports, Racing, Wrestling \\
Miscellaneous & Simulation, Flight, Party, Real-Time \\
\hline
\end{tabular}
\vspace{0.5cm}
\caption[temp]{
\label{table:genreclasses}
Our proposed grouping of genres into 5 classes based on similarity
}
\end{table}

{\bf Game scores} -
We define the G-Score of a game as an average of critic and user ratings, details are specified in Section \ref{sec:scorepred}. We observe that most games have G-Scores above 40 and only a small fraction of games have a G-Score below 40 as shown in Table \ref{table:dataset}. This results in some inter-class bias. The main reason for this is that most video games that would potentially have a bad G-Score would either not have trailers or not have any associated critic/user ratings since most people would not play the game in the first place and hence would not be present in the dataset.

\begin{figure*}[ht]
    \begin{minipage}[b]{1\linewidth}
           \includegraphics[width=1.03\textwidth]{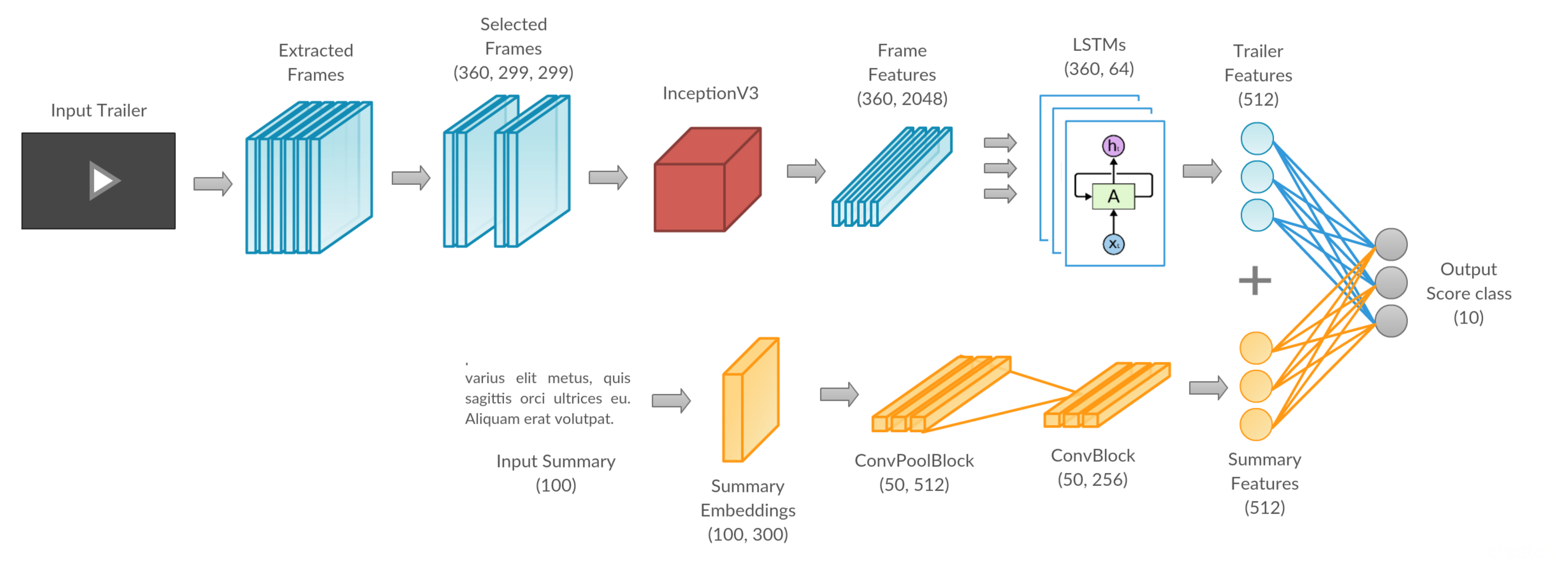}
           \caption{An overview of our pipeline corresponding to Model-1. We start with the trailers and summaries as inputs and predict the G-Score classes as outputs. The Inception-V3 pre-trained model is used to extract features of video frames. ConvPoolBlocks and ConvBlocks are described in Figure \ref{fig:convblocks}. The output sizes at each of the layers are mentioned in the figure.}
        \label{fig:overview}
    \end{minipage}
\end{figure*}

\section{Score Prediction}
\label{sec:scorepred}
Each video game has a user rating $R_u$ and critic rating $R_c$ associated with it. We define the G-Score of a game (S) as follows,
\begin{equation}
S = \frac{R_u + R_c}{2}
\label{equation:score}
\end{equation}
and aim to predict this G-Score. The G-Score essentially represents how good a game is. 

Critic ratings are collected from a large number of critics and a weighted average is computed to form the final critic rating $R_c$. The weights of individual critics depend on the overall stature of the critic. Formally, if $R_c^i$ corresponds to the rating of critic $i$ and ${\alpha}^i$ is the weight associated with critic $i$ and there are $M$ critics then,
\[ R_c = \sum_{i=1}^{M} {\alpha}^i * R_c^i \]

The number of critics that review games vary from game to game since popular games often get a larger number of ratings as compared to others that are not so popular. Critic weights ${\alpha}^i$ are based on how well the critics performed in the past (well written, insightful reviews etc). This is determined by Metacritic staff who handle the website from where we collected our data.

The user rating $R_u$ is computed as an average of all user ratings submitted for the game. Formally, if $R_u^i$ corresponds to the rating of user $i$ and there are $N$ users then,
\[ R_u = \sum_{i=1}^{N} R_u^i \]

Regardless of the number of users and critics that rate a game, we compute the final score as mentioned in Equation \ref{equation:score}. We consider both trailers and summaries as inputs in order to predict this G-Score using our proposed model. We quantize the G-Scores to $10$ classes since predicting the G-Score directly is a regression problem which is harder to tackle compared to classification problems.

\subsection{Trailers}
Each video game is associated with a trailer that we use in order to predict the G-Score.

{\bf  Trailer frame selection} -
Since videos are captured with a frame-rate of 24 fps, it is infeasible to use them as they are since the sheer number of frames are too many. Hence, we propose a method to pick frames in a certain manner that would allow us to maximize the information we obtain from game trailers. Firstly, we reduce the frame-rate to 4 fps while extracting the frames from the video. We then follow the frame selection algorithm mentioned in Algorithm \ref{alg:frameselection} in order to select frames. This allows us to capture important information at various parts of trailers. The reason we skip frames is that most trailers have a sequence of events that go on for a while before transitioning to the next sequence of events. Upon observation, we use a skip of 150 frames as a good approximation. We skip the first 50 frames since most trailers have textual information during the start of the video such as the developer titles, age ratings etc.

\begin{algorithm}[h]
{
  \caption{Frame selection for trailers:}
  \label{alg:frameselection}       
  \begin{algorithmic}[1]
  \State Consider we have a set of N frames $F_{1}, F_{2}, ..., F_{N}$
  \State $f_{start}$ = 50
  \While{$f_{start} < N$}
  \For{$j = 0$, $j{+}{+}$, while $j < 10$}
  \If{$f_{start} + j <= N$}
  \State Select frame $F_{{f_{start} + j}}$
  \Else
  \State Break
  \EndIf
  \EndFor
  \State $f_{start} += 150$
  \EndWhile
  \end{algorithmic}
  }
\end{algorithm}

{\bf  Trailer features} -
We use the pre-trained Inception-V3 \cite{DBLP:journals/corr/SzegedyVISW15} model to extract features from each of the frames selected in the previous step. The model was pre-trained on ImageNet \cite{imagenet_cvpr09} and hence generalizes well to a wide range of images. We extract the features of the \textit{Avg\_Pool} layer (the penultimate layer in the network) which gives us a feature representation of 2048 elements per frame. Considering all the frames, we get a vector having dimensions $(M, 2048)$ where $M$ is the number of frames we selected by the frame selection algorithm \ref{alg:frameselection} as our final trailer features.

\begin{figure}[h]
    \begin{minipage}[b]{1\linewidth}
           \includegraphics[width=1\textwidth]{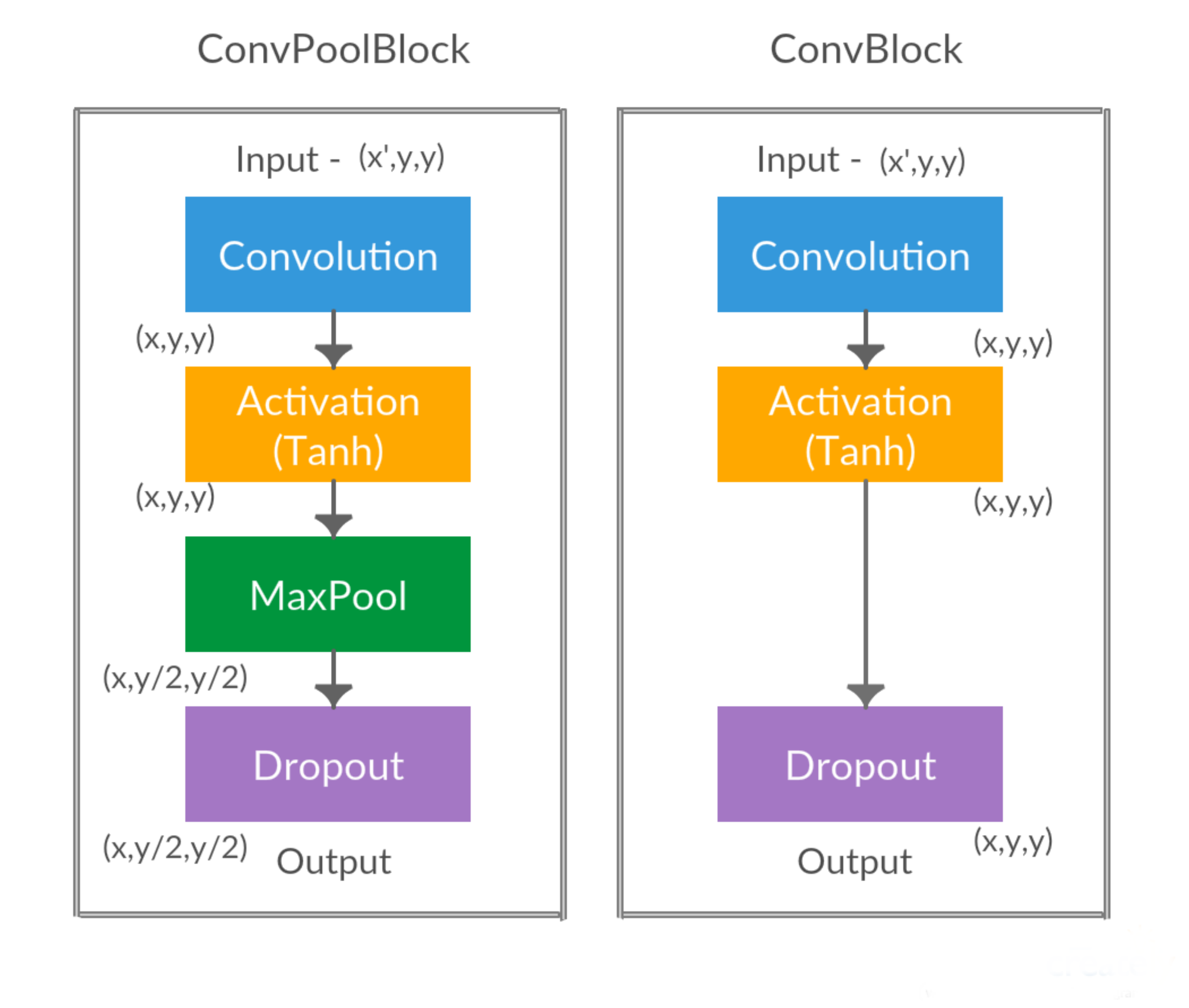}
           \caption{ConvPoolBlocks and ConvBlocks are used in order to process the summaries in our proposed models. A ConvPoolBlock consist of Convolution, Tanh, MaxPool and Dropout layers. A ConvBlock consists of Convolution, Tanh and Dropout layers.}
        \label{fig:convblocks}
    \end{minipage}
\end{figure}

\subsection{Summaries}
Considering all the summaries we have, we create a dictionary where each word is given an index. We then go through each of the summaries replacing words with their corresponding indexes. Finally, we resize the summaries to a size of 100 by trimming the summaries if they are larger and padding them with zeros if they are smaller.

\begin{figure*}[ht]
    \begin{minipage}[b]{0.5\linewidth}
           \includegraphics[width=\textwidth]{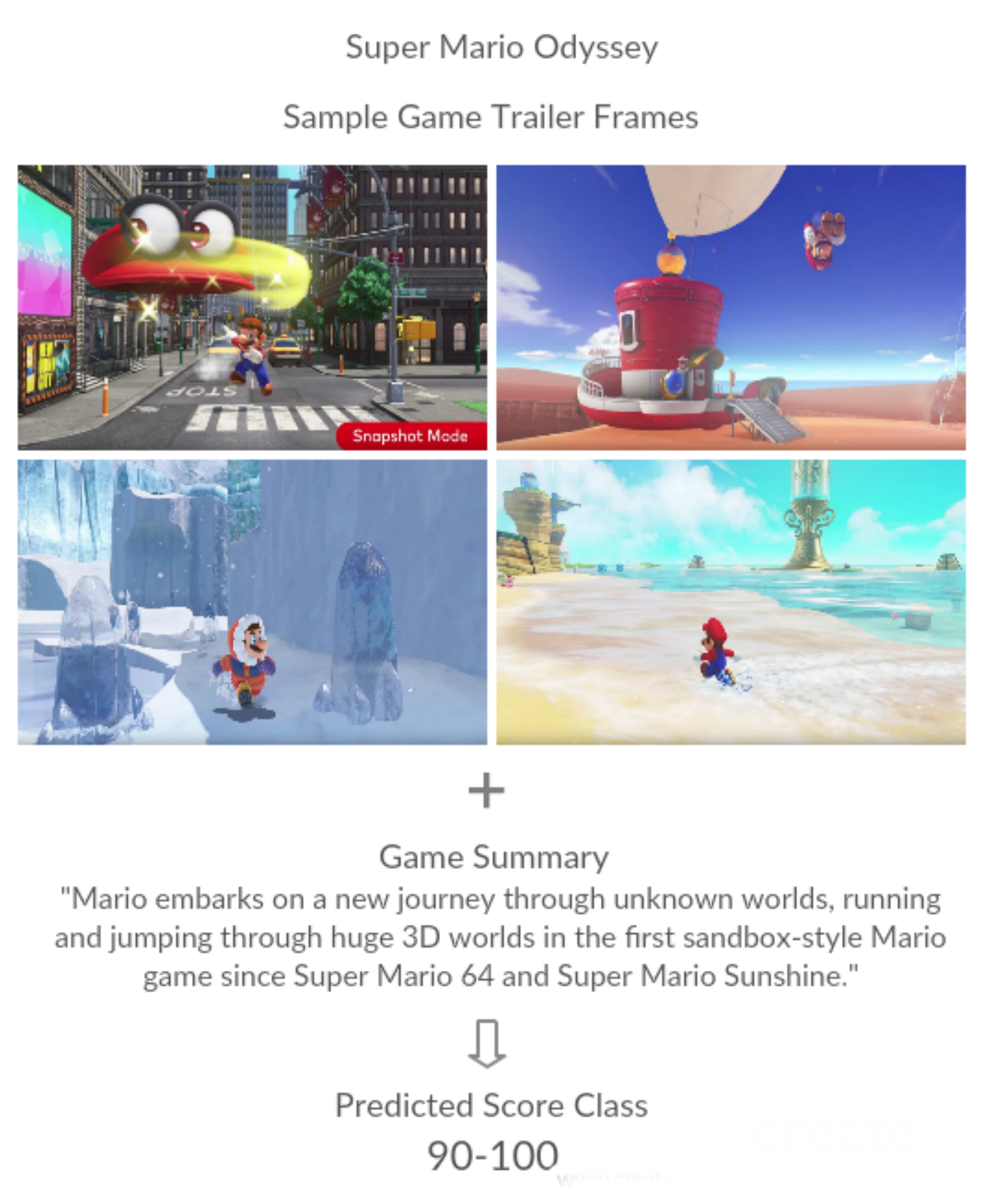}
    \end{minipage}
    {\vrule width 0.1pt}
    \begin{minipage}[b]{0.5\linewidth}
           \includegraphics[width=\textwidth]{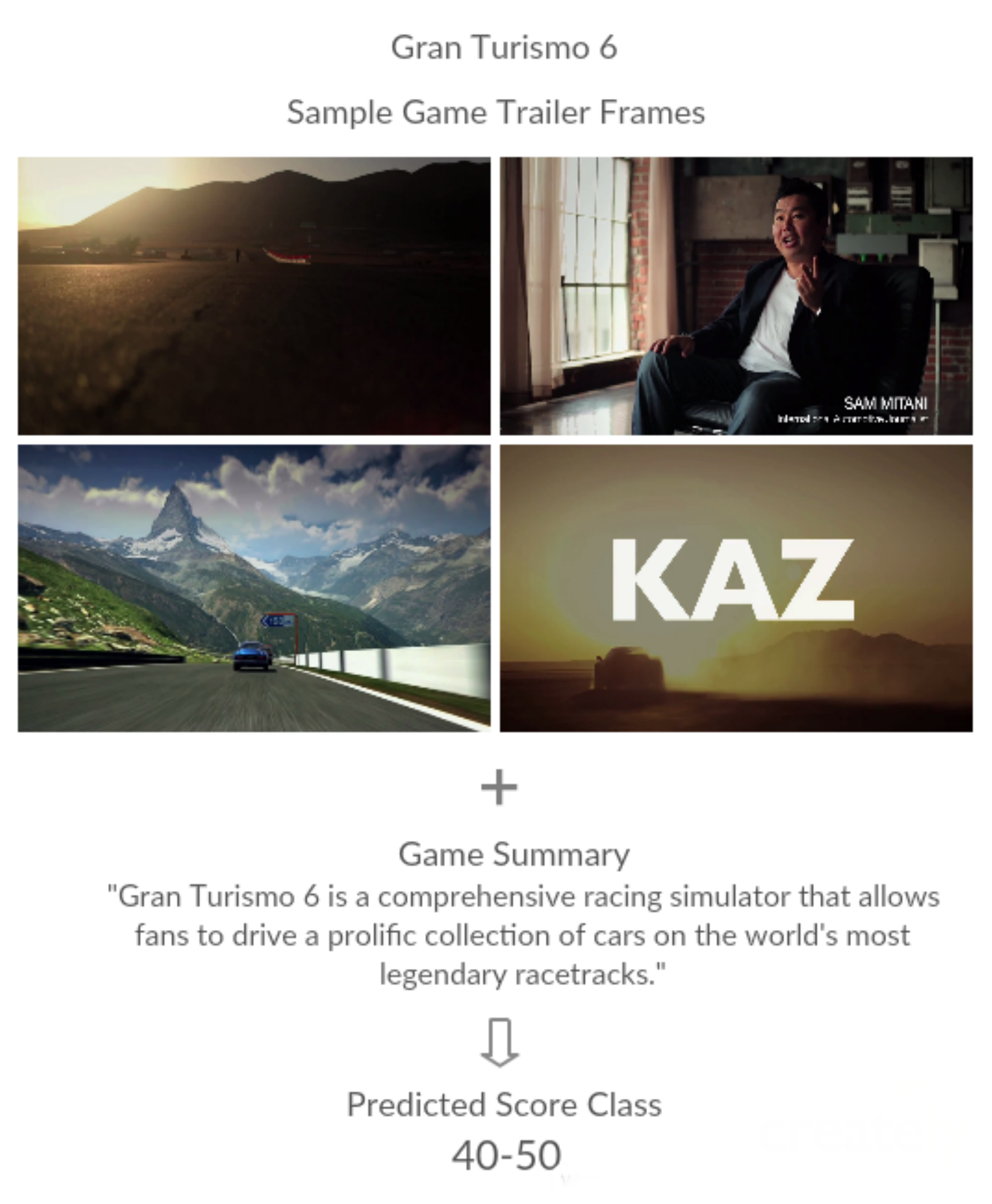}
    \end{minipage}
    \caption{Qualitative examples where the model predicts the correct G-Score class for Super Mario Odyssey (left) and the incorrect G-Score class for Gran Turismo 6 (right). The true G-Score for Super Mario Odyssey is 93 and the true G-Score for Gran Turismo 6 is 81.}
    \label{fig:qualitative}
\end{figure*}

\subsection{Method}
{\bf Model-1} -
We propose a deep learning based architecture as outlined in Figure \ref{fig:overview} that uses a combination of recurrent and convolution based networks which allows us to process both trailer features and summaries in order to predict the G-Score of a video game. The frame features are fed to multiple levels of LSTMs\cite{Hochreiter:1997:LSM:1246443.1246450} that finally output a vector of size $512$. The summaries are fed to an embedding layer that dynamically generates embeddings having a size of $300$. These embeddings are then fed to a convolution-based network as depicted in Figures \ref{fig:overview} and \ref{fig:convblocks} that outputs a vector of size $512$. Finally, these vectors are concatenated and passed along to a linear layer that outputs the G-Score class. We also perform experiments on multiple other model architectures, however, this gives us the best results.

{\bf Model-2} -
We use a time distributed CNN over extracted frame features to generate a small embedding for each frame which are then concatenated and fed to a fully connected layer in order to produce the final output vector. Summaries are passed through a CNN, similar to what was done earlier. The outputs from both the LSTM and CNN are concatenated and a linear layer is applied to predict the final output. One significant advantage of this approach was that the model had a very small number of parameters since the time distributed CNN shares weights across time. This would be an ideal model to use in memory constrained scenarios such as mobile computing.

{\bf Model-3} -
We use a 3D CNN \cite{Ji:2013:CNN:2412386.2412939} over the frames to generate an output embedding for the trailer. Summaries are passed through a CNN, similar to what was done earlier. The outputs are then concatenated and passed to a linear layer to predict the final G-Score class.

We also tried generating sentence embeddings using Doc2Vec \cite{DBLP:journals/corr/LauB16} for each of the summaries but they didn't give us the best results and hence we stuck to dynamic embeddings as mentioned earlier. The three models mentioned here consider both the trailer and summary as inputs. In order to validate our claim that the use of summaries gives us accuracy improvements, we also perform the same experiments without considering summaries on each of the proposed models and report accuracies in Table \ref{Accuracies-Rating2}. This shows that using summaries along with trailers gives us significant improvements of over 5\%.

\subsection{Implementation Details}
We implemented the proposed models in the Keras \cite{chollet2015keras} framework over the Tensorflow \cite{tensorflow2015-whitepaper} backend. We use the Cross-entropy loss at the output of our model and use the Adam optimizer with a learning rate of 1e-4 and a decay of 1e-6 in order to train the model. We use \textit{tanh} activations instead of ReLU throughout the model as it helps us achieve better accuracies. We also include multiple Dropout \cite{srivastava2014dropout} layers to allow the model to generalize well.

To evaluate our model, we perform 10-fold cross-validation and provide results. Further details on Model-1 (our best model) can be found in the code submitted along with this work at \href{https://goo.gl/fYiEfq}{https://goo.gl/fYiEfq}.

\section{Results}
On each of our proposed models, we perform 10-fold cross-validation and consider the mean as our final accuracy. We observe a significant increase in accuracy with the inclusion of summaries as inputs along with the trailers. Model-1 gives us the best results in terms of accuracy. We believe the main reason for this is that Inception-V3 is trained on ImageNet which is a huge dataset of more than 1M images. Hence, it provides us with feature representations that are rich and meaningful.

\begin{table}[h]
\centering
\setlength{\extrarowheight}{0.5em}
\begin{tabular}{|c|c|c|}
\hline Model & Input & Accuracy \\ \hline
\multirow{3}{*}{Model-1} & Trailer Only & 65.2 \\
 & \bf{Trailer and Summary} & \bf{70.5} \\
 & Improvement & \textcolor{green}{+5.3\%} \\
\hline
\multirow{3}{*}{Model-2} & Trailer Only & 63.3 \\
 & Trailer and Summary & 66.6 \\
 & Improvement & \textcolor{green}{+3.3\%} \\
\hline
\multirow{3}{*}{Model-3} & Trailer Only & 64.5 \\
 & Trailer and Summary & 68.8 \\
 & Improvement & \textcolor{green}{+4.3\%} \\
\hline
\end{tabular}
\vspace{0.5cm}
\caption[temp]{
\label{Accuracies-Rating2}
Results on predicting G-Scores using 10-fold cross-validation. We present mean accuracies on all three of our models considering trailers only as inputs as well as both trailers and summaries as inputs. Improvements obtained with each of the models are also mentioned.
}
\end{table}

Model-2 has a very small number of parameters which is why it is well suited for use in portable devices and memory constrained situations such as mobile processing. This, however, comes at a cost that the accuracy is lower than Model-1.

\subsection{Qualitative Analysis}
A few qualitative results have been provided where the network performs well in one case but fails at the other as shown in Figure \ref{fig:qualitative}. Gran Turismo 6 has a true G-Score of 81 but we predict a G-Score class of 40-50. The main reason this fails is that the trailer has multiple overlay texts and game-play irrelevant clips. A simple solution to frames containing overlay text is to ignore them before feeding them to model. We could also process these frames separately, extracting the text from them and using them as inputs along with the summaries of games. Since non game-play scenes do not contribute any significant information when scoring a game, the model would misinterpret this information hence resulting in incorrect G-Scores. In the example provided, refer Figure \ref{fig:qualitative}, the frame containing the person would get a feature representation from Inception-V3 that has no relevance to the game and would ultimately contribute to noise. Handling non game-play scenes in trailers is an issue that is hard to tackle and is one of the shortcomings of this work. An approach towards this would be to train a model that takes a frame as an input and predicts if the frame is a game-play scene or not given the video as a reference.

\subsection{Empirical Validation}
We validate our claim that summaries provide information that is quite useful while predicting the G-Score of a video game. Hence, using both the trailers and summaries allows us to predict with a much better accuracy. We conduct a significance test where we perform experiments on predicting the G-Score of a video game based on the trailer alone and show that we gain significant accuracy improvements of over $5\%$ when we use both the summaries and trailers in order to predict the G-Score as mentioned in Table \ref{Accuracies-Rating2}. Most of the times, we have the summary at our disposal along with the trailers of games and hence, using information from multiple modalities helps us develop models that perform better.

\section{Conclusion and Future Work}
In this work, we show how valuable multimodal knowledge is at performing a task at hand. In most real-life scenarios we would have multimodal information available which could be utilized to train better models. We also provide a new VGD dataset that is a dataset on video games, a first of its kind. We propose multiple models that work under different scenarios such as memory constrained settings etc. We plan to apply our approach to movie trailers and summaries in order to show the generalizability of our approach. We plan to take care of overlay texts that occur in trailers by processing them separately in order to produce better results. Finally, we also plan to include audio in order to improve our prediction accuracies.

\bibliographystyle{IEEEtran}
\bibliography{IEEEabrv}

\end{document}